\pdfoutput=1

\documentclass[11pt]{article}

\usepackage[review]{emnlp2021}

\usepackage{times}
\usepackage{latexsym}
\usepackage{graphicx}
\usepackage[T1]{fontenc}

\usepackage[utf8]{inputenc}

\usepackage{microtype}
\usepackage{amsmath}
\usepackage{amssymb}

\title{Counterfactual Multihop QA: A Cause-Effect Approach for Reducing Disconnected Reasoning}

\author{Wangzhen Guo  \quad Qinkang Gong \quad  Hanjiang Lai\thanks{*Corresponding Author} \\
  Sun Yat-Sen University \\
  \texttt{\{guowzh6,gongqk\}@mail2.sysu.edu.cn} \quad \texttt{laihanj3@mail.sysu.edu.cn}
}

\begin{document}
\maketitle
\begin{abstract}
Multi-hop QA requires reasoning over multiple supporting facts to answer the question. However, the existing QA models always rely on shortcuts, e.g., providing the true answer by only one fact, rather than multi-hop reasoning, which is referred as \textit{disconnected reasoning} problem. To alleviate this issue, we propose a novel counterfactual multihop QA, a causal-effect approach that enables to reduce the disconnected reasoning. It builds upon explicitly modeling of causality: 1) the direct causal effects of disconnected reasoning and 2) the causal effect of true multi-hop reasoning from the total causal effect. With the causal graph,  a counterfactual inference is proposed  to disentangle the disconnected reasoning from the total causal effect, which provides us a new perspective and technology to learn a QA model that exploits the true multi-hop reasoning instead of shortcuts. Extensive experiments have conducted on the benchmark HotpotQA dataset, which demonstrate that the proposed method can achieve notable improvement on reducing disconnected reasoning. For example, our method achieves 5.8\% higher points of its Supp$_s$ score on HotpotQA through true multihop reasoning. The code is available at supplementary material.
\end{abstract}

\section{Introduction}

Multi-hop question answering (QA)~\cite{simplepipeline,CogQA,pathRetriver,graph-necessity} requires the model to reason over multiple supporting facts to correctly answer a complex question. It is a challenging task, and many datasets, e.g., HotpotQA~\cite{hotpotqa} and approaches~\cite{HGN,ASIO} have been proposed  for this reasoning task. 

\begin{figure*}[t]
    \centering
    \includegraphics[width=14cm]{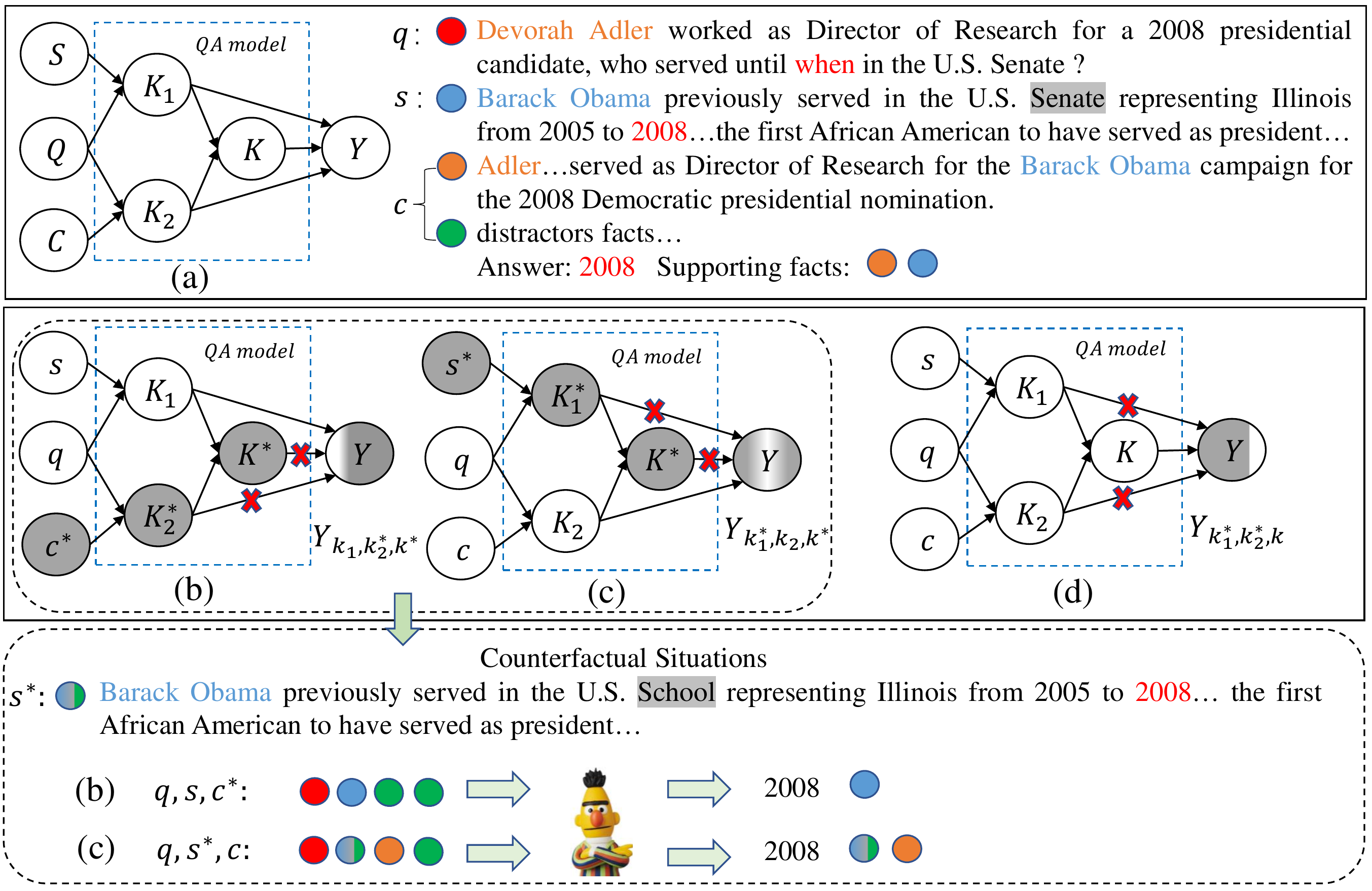}
    \caption{Illustration of disconnected reasoning in multi-hop QA, 
    where red node denotes question $Q$, blue node is a supporting fact $S$, and orange and green nodes denote the remaining facts $C$. Deep gray nodes mean their variables are reference values instead of the given values, (e.g., $S=s^*$ instead of $S=s$). \textbf{(a):} Causal graph of multi-hop QA model;  \textbf{(b):} is a possible scenario of disconnected reasoning, which uses only one fact $s$ to answer the question. \textbf{(c):} is another possibility of disconnected reasoning, e.g., the exclusive method to find whether $s$ is a supporting fact by a process of elimination of other facts $c$.
    \textbf{(d):} is the true multi-hop reasoning. All facts $s$ and $c$ are taken into considered to produce the answer. 
    }
    \label{fig:causal_graph}
\end{figure*}

One of the main problems of multihop QA models is \textit{disconnected reasoning}~\cite{dire}, which allows the models to exploit the reasoning shortcuts \citep{Advexamples,robustifying} instead of multi-hop reasoning to cheat and obtain the right answer. Taking Fig.~\ref{fig:causal_graph} as an example, to answer the question \textit{``until when in the U.S. Senate"}, we should consider two supporting facts to infer the answer ``Devorah Adler $\stackrel{Director \ of \ Research \ for}{\Longrightarrow}$ Barack Obama $\stackrel{served \ from \ 2005 \ to \ 2008}{\Longrightarrow}$ 2008". However, one may also infer the correct answer by just utilizing the types of problems, e.g., we can find the corresponding fact \textit{``from 2005 to 2008"} in the contexts without reasoning to answer this type of question \textit{``until when"}. 

One possible solution for reducing the disconnected reasoning is to strengthen the training dataset via extra annotations or adversarial examples, which make it cannot find the correct answers by only one supporting fact. For example, 
\citet{Advexamples} constructed the adversarial examples to generate better distractor facts. \citep{dire} firstly defined a evaluate measure, \textbf{DiRe} in short, to measure how much the QA model can cheat via disconnected reasoning. Then, a transformed dataset is constructed to reduce disconnected reasoning.  Besides, counterfactual intervention \citep{robustifying,connecting} had also been explored to change the distribution of the training dataset. These methods improve the generalizability and interpretability of the multi-hop reasoning QA model via balancing the train data, which is noted as debiased training in QA model~\citep{counterfactualVQA}. However, when the existing approaches decrease the disconnected reasoning, the original performance also drops significantly. It is still challenging to reduce disconnected reasoning while maintaining the same accuracy on the original test set. 



Motivated by causal inference \citep{theBookOfWhy,pearl2022direct,counterfactualVQA}, we utilize the counterfactual reasoning to reduce the disconnected reasoning in multi-hop QA and also obtain the robust performance on the original dataset. 
We formalize a causal graph to reflect the causal relationships between question ($Q$), contexts and answer ($Y$). To evaluate the  disconnected reasoning, contexts are further divided into two subsets: $S$ is a supporting fact and $C$ are the remaining supporting facts. Hence, we can formulate the disconnected reasoning as two natural direct causal effects of $(Q,S)$ and $(Q,C)$ on $Y$ as shown in Fig.~\ref{fig:causal_graph}. With the proposed causal graph, we can relieve the disconnected reasoning by disentangling the two natural direct effects and the true multi-hop reasoning from the total causal effect. A novel counterfactual multihop QA is proposed to disentangle them from the total causal effect. 
We utilize the generated probing dataset proposed by \citep{dire} and DiRe to measures how much the proposed multi-hop QA model can reduce the disconnected reasoning. Experiment results show that our approach can substantially decrease the disconnected reasoning while guarantee the strong performance on the original test set. The results indicate that the proposed approach can reduce the disconnected reasoning and improve the true multi-hop reasoning capability.

The main contribution of this paper is threefold. Firstly, our counterfactual multi-hop QA model formulates disconnected reasoning as two direct causal effects on answer, which is a new perspective and technology to learn the true multi-hop reasoning. Secondly, our approach achieves notable improvement on reducing disconnected reasoning compared to various state-of-the-arts. Thirdly, our causal-effect approach is model-agnostic and can be used for reducing disconnected reasoning in many multi-hop QA architectures.



\section{Related Work}
Multi-hop question answering (QA) requires the model to retrieve the supporting facts to predict the answer.  Many approaches and datasets have been proposed to train QA systems. For example, HotpotQA~\citep{hotpotqa} dataset is a widely used dataset for multi-hop QA, which consists of fullwiki setting~\citep{Entity-centric,Semantic,query-iter,sentence+rnn,hopretriever,MDPR} and distractor setting~\citep{Decomposition,extraction,DFGN,Advexamples,dire}.

In fullwiki setting, it firstly finds relevant facts from all Wikipedia articles, and then finish the multi-hop QA task with the found facts. The retrieval model is important in this setting. For instance, SMRS~\citep{Semantic} and DPR~\citep{DPR} found the implicit importance of retrieving relevant information in the semantic space.  Entity-centric~\citep{Entity-centric},  CogQA~\citep{CogQA} and Golden Retriever~\citep{query-iter} explicitly used the entity that is mentioned or reformed in query key words to retrieve next hop document. Furthermore, PathRetriever~\citep{pathRetriver} and HopRetriever~\citep{hopretriever} can iteratively select the documents to form a paragraph-level reason path using RNN. MDPR~\citep{MDPR} retrieved passages only using dense query vector in many times. These methods hardly discuss the QA model's disconnected reasoning problem. 

In distractor setting, 10 paragraphs, two gold paragraphs and eight distractors, are given.  Many methods have been proposed to strengthen the model's capability of multi-hop reasoning, using graph neural network~\citep{DFGN,HGN,graph-necessity} or adversarial examples or counterfactual examples~\citep{Advexamples,robustifying} or the sufficiency of the supporting evidences~\citep{dire} or make use of the pretrained language models~\citep{transformerxh,bigbird}. 

However, \citet{mpreason-analysis} demonstrated that many compositional questions in HotpotQA can be answered with a single hop. It means that  QA models can take shortcuts instead of multi-hop reasoning to produce the corrected answer. To relieve the issue, \citet{Advexamples} added adversarial examples as hard distractors during training. Recently, \citep{dire} proposed an approach, DiRe, to measure the model's disconnected reasoning behavior and use the supporting sufficiency label to reduce the disconnected reasoning. \citet{robustifying} selected the supporting evidence according to the sentence causality to the predicted answer, which guarantees the explainability of the behavior of the model. While, the original performance also drops when reducing the disconnected reasoning. 

\textbf{Causal Inference.} Recently, causal inference \citep{theBookOfWhy,pearl2022direct} has been applied to many tasks of natural language processing, and it shows promising results and provides strong interpretability and generalizability. The representative works include counterfactual intervention for visual attention~\citep{cal}, causal effect disentanglement for VQA~\citep{counterfactualVQA}, the back-door and front-door adjustments~\citep{cross-back-door1,causal-back-door2,causal-front-door}. Our method can be viewed as a complement of the recent approaches that utilize the counterfactual inference \citep{robustifying,connecting} to identify the supporting facts and predict answer.

\section{Preliminaries} \label{Preliminaries}






In this section, we use the theory of causal inference \citep{theBookOfWhy,pearl2022direct} to formalize our multi-hop reasoning method. Support that we have a multi-hop dataset $D$ and each instance has the form of $(Q,P;Y)$, where $Q$ is a question and $P=\{s_1, s_2, \cdots, s_n \}$ is a context consisting of a set of $n$ paragraphs. And $Y$ is the ground-truth label. Given a question $Q$ with multiple paragraphs as a context $P$, the multi-hop QA models are required to identify which paragraphs are the supporting facts and predict an answer using the supporting facts.

\textbf{Causal graph}. In multi-hop QA, multiple supporting facts are required to predict the answer. While QA models may use only one fact to give the answer, which is referred as disconnected reasoning. For example, given a question $q$ and only a paragraph $s$, the disconnected reasoning model can predict the correct answer or correctly determine whether the paragraph $s$ is the supporting fact. Hence, to define the causal graph of disconnected reasoning, the context $P$ is further divided into a paragraph $S$ and the remaining paragraphs $C$. Now the $(Q,P;Y)$ becomes $(Q,S,C;Y)$. That is each instance $(Q,P;Y)$ is converted into $n$ examples, i.e., $(q,s_1,C=\{s_2,\cdots,s_n\};Y(s_1))$, $\cdots$, $(q,s_n,C=\{s_1,\cdots,s_{n-1}\};Y(s_n))$, where $Y(s_i)=\{F_{s_i};A\}$ includes the supporting fact and answer, where $A$ is the answer and $F_{s_i}=1$ means the paragraph $s$ is the supporting fact otherwise it is not. For each example, we consider the disconnected reasoning with the  fact $S$ (not $C$). 

The causal graph for multi-hop QA is shown in Figure~\ref{fig:causal_graph} (a), where nodes denote the variables and directed edges represents the causal-and-effect relationships between variables. The paths in Figure~\ref{fig:causal_graph} (a) are as follows. 

$(Q,S) \to K_1 \to Y$: $(Q,S) \to K_1$ denotes that the feature/knowledge $K_1$ is extracted from the question $(Q)$ and the paragraph $(S)$ via the QA model backbone, e.g., BERT. $K_1 \to Y$ represents the process that the label $Y$ is predicted by only using the $K_1$.  

$(Q,C) \to K_2 \to Y$: Similarly, the feature/knowledge $K_2$ extracted from the question $(Q)$ and the remaining paragraphs $(C)$ is used to predict the label $Y$. 

$(Q,S,C) \to (K_1,K_2) \to K \to Y$: This path indicates that the QA model predicts the label $Y$ based on both the $K_1$ and $K_2$.

Based on the above, the effect of $Q,S,C$ on Y can be divide into: 1) shortcut impacts, e.g., $(Q,S) \to K_1 \to Y$ and $(Q,C) \to K_2 \to Y$, and 2) reasoning impact, e.g., $(Q,S,C) \to (K_1,K_2) \to K \to Y$. The shortcut impacts capture the direct effect of $(Q,S)$ or $(Q,C)$ on $Y$ via $K_1 \rightarrow Y$ or $K_2 \rightarrow Y$. The reasoning impact captures the indirect effect of $(Q,S,C)$ on $Y$ via $K \rightarrow Y$. 

Hence, to reduce the multi-hop QA model's disconnected reasoning proposed in \citep{dire}, we should exclude shortcut impacts ($ K_1 \rightarrow Y$ and $K_2 \rightarrow Y$) from the total effect.


\textbf{Counterfactual definitions}.
Figure~\ref{fig:causal_graph} (a) shows the causal graph. From causal graph to formula, we denote the value of $Y$, i.e., the answer $A$ (e.g., 2008) or the supporting paragraph $F_{s}$ (e.g., is the paragraph $s$ supporting fact?), would be obtained when question $Q$ is set to $q$, the paragraph $S$ is set to $s$ and the remaining paragraphs $c=P-s$ are used as the context $C$, which is defined as 
$$
\begin{aligned}
&Y_{q,s,c}(A)=Y(A\ |\ Q=q,S=s,C=c), \\ 
&Y_{q,s,c}(s)=Y(s\ |\ Q=q,S=s,C=c).
\end{aligned}
$$
For simplicity, we omit $A,s$ and unify both equations as $Y_{q,s,c}=Y_{q,s,c}(A)$ or $Y_{q,s,c}=Y_{q,s,c}(s)$. Since the causal effect of $q,s,c$ on $Y$ via $K_1,K_2,K$ on $Y$, we have 
\begin{equation}
    Y_{k_1,k_2,k}=Y_{q,s,c}
\end{equation}
in the following discussion. 

To disentangle the shortcut impacts from the total causal effect, we use the counterfactual causal inference to block other effects. To model the $K_1 \to Y$, the counterfactual formulation
\begin{equation}
   K_1 \to Y : Y_{k_1,k_2^*,k^*},
    \label{nde1}
\end{equation}
 which describes the situation where $K_1$ is set to the original value $k_1$ and $K$ and $K_2$ are blocked. The $k^*$ and $k^*_2$ are the counterfactual notations. The $k_1$ and $k^*,k^*_2$ represent the the two situations where the $k_1$ is under treatment in the factual scenario and $k^*,k^*_2$ are not under treatment~\cite{theBookOfWhy} in the counterfactual scenario.  The same definitions for other two effects as
 \begin{equation}
   K_2 \to Y : Y_{k_1^*,k_2,k^*},
   \label{nde2}
\end{equation}
and 
\begin{equation}
   K \to Y : Y_{k_1^*,k_2^*,k}.
\end{equation}


\textbf{Causal effects}. According the counterfactual definitions, total effect of $q,s,c$ on $Y$ can be decomposed into the natural direct effects of $K_1,K_2$ on $Y$ and the effect of $K$ on $Y$ as discussed before. The two natural direct effects cause the disconnected reasoning problem, and the effect of $K \to Y$ is the desired multi-hop reasoning. 

As shown in Figure~\ref{fig:causal_graph} (b), the effect of $K_1$ on $Y$ with $K_2,K$ blocked and the effect of $K_2$ on $Y$ with $K_1,K$ blocked can be easily obtained by setting the $S$ or $C$ to counterfactual values (please refer to Section~\ref{implementation} for more details). While the effect of $K$ on $Y$ can not be obtained by changing the values $S$/$C$ to $S^*$/$C^*$. We follow ~\cite{counterfactualVQA} and total indirect effect (TIE) is used to express the effect of $K$ on $Y$, which is formulated as
\begin{equation}
    \label{tie}
    Y_{k^*_1,k^*_2,k} = Y_{k_1,k_2,k} - Y_{k_1,k_2,k^*}.
\end{equation}

\section{Counterfactual Multihop QA}\label{implementation}

Following the former formulations, we propose to construct the counterfactual examples to estimate the natural direct effect of $K_1$ and $K_2$, as well as using parameters to estimate the total indirect effect of $K$. And our calculation of $Y$ in Eq. (\ref{nde1}), (\ref{nde2}) and (\ref{tie}) is parametrized by a neural multi-hop QA model $\mathcal{F}$. Please note that $\mathcal{F}$ can be any multi-hop QA models and our method is model-agnostic.

\subsection{Disentanglement of causal effect}
\label{estimationOfeffect}
$K_1 \to Y$. Specifically, in Eq.~(\ref{nde1}), the $Y_{k_1,k_2^*,k^*}$ describes the situation where $K_1$ is set to the factual value with $Q=q, S=s$ as inputs, and $K_2$/$K$ are set to the counterfactual values. Taking Figure~\ref{fig:causal_graph} as an example, the QA model only consider the interaction between question $q$ and a given paragraph $s$. 
The remaining paragraphs $c$ are not given. It is the disconnected reasoning~\citep{dire}. To obtain the counterfactual values of $K_2$ and $K$, we can set the context $C$ as its counterfactual sample, and we have
\begin{equation}
    Y_{k_1,k_2^*,k^*} = Y_{q,s,c^*} = \mathcal{F}(q,s,c^*).
\end{equation}
In this paper, we randomly sample the remaining contexts from the training set to construct the counterfactual $c^*$. It represents that the no-treatment or the prior knowledge of the remaining context is unknown. In the implementation, we randomly sample the remaining contexts in a mini-batch to replace the $c$ in original triple example ($q,s,c$) and obtain the corresponding counterfactual triple example ($q,s,c^*$). With that, we can feed it into the QA model to get $Y_{q,s,c^*}$.

$K_2 \to Y$. Similarly, in Eq.~(\ref{nde2}), the $Y_{k_1^*,k_2,k^*}$ describes the situation where $K_2$ is set to the factual value with the inputs $Q=q$ and $C=c$. The $K_1$ and $K$ are set to the counterfactual values with the counterfactual sample $S=s^*$ as input, which is defined as
\begin{equation}
    Y_{k_1^*,k_2,k^*} = Y_{q,s^*,c} = \mathcal{F}(q,s^*,c).
    \label{equlll}
\end{equation}

One may argue that how to predict the label when $S$ is set to the counterfactual values? As the example shown in Figure~\ref{fig:causal_graph}, even without the paragraph $s$ as input, the QA model still can infer the paragraph $s$ is supporting fact via wrong reasoning: since all paragraphs in $C$ do not include the words about time range, the rest paragraph $s$ should be the supporting fact to answer the ``until when" question. It is the exclusive method. The wrong reasoning is caused by an incorrect interaction between the paragraph $s$ and the remaining context $c$. Hence, the $S$ is set to the counterfactual values can correct such incorrect interactions. 

Hence, in the implementation, we use the adversarial examples as suggested in~\cite{Advexamples} to construct the counterfactual $s^*$, which aims to remove the incorrect interaction and perform multi-hop reasoning. 
Specifically, we randomly perturb the 15\% tokens of the paragraph $s$, 80\% of which will be replaced by other random tokens, 10\% of which will be replaced by the mask token (e.g. $[MASK]$) of the tokenizer, and 10\% of which will keep unchanged. 
After that, we obtain another counterfactual triple example ($q,s^*,c$), we can feed it into QA model to get $Y_{q,s^*,c}$.

$K \to Y$.  In Eq.~(\ref{tie}), $Y_{q,s,c}$ indicates that the question $s$, paragraph $s$ and remaining context $c$ are visible to the QA model $\mathcal{F}$:
\begin{equation}
    Y_{k_1,k_2,k} = Y_{q,s,c} = \mathcal{F}(q,s,c).
\end{equation}

The main problem is that $Y_{k_1,k_2,k^*}$ is unknown. It is also hard to use the counterfactual samples of $Q,S,C$ to obtain its value since the $q,s$ and $c$ should be the factual values for $k_1, k_2$. In this paper, we follow the work~\cite{counterfactualVQA} and also assume the model will guess the output probability under the no-treatment condition of $k^*$, which is represented as
\begin{equation}
    Y_{k_1,k_2,k^*} = \mathcal{C},
\end{equation}
where $\mathcal{C}$ is the output and a learnable parameter. 
Similar to Counterfactual VQA~\citep{counterfactualVQA}, we guarantee a safe estimation of $Y_{k_1,k_2,k^*}$ in expectation. 

\textbf{Training objective}. From the above discussion, we can estimate the total causal effect:
\begin{equation}
\label{newY}
    \begin{aligned}
        Y \leftarrow Y_{q,s,c^*} + Y_{q,s^*,c} + Y_{q,s,c} - \mathcal{C}.
    \end{aligned}
\end{equation}
Note that $Y$ can be any ground-truth labels of answer span prediction, supporting facts identification or answer type prediction. See Appendix \ref{implementaionDetails} for further implementation details.

\subsection{Training and Inference}

\begin{table*}[ht]
\centering
\begin{tabular}{ccccccccccc}
\hline
 & \multicolumn{2}{c}{\textbf{$\rm Ans$}} & \multicolumn{2}{c}{\textbf{$\rm Supp_p$}} & \multicolumn{2}{c}{\textbf{$\rm Supp_s$}} & \multicolumn{2}{c}{\textbf{$\rm Ans+Supp_p$}} & \multicolumn{2}{c}{\textbf{$\rm Ans+Supp_s$}}\\
& original & dire\text{$\downarrow$} & original & dire\text{$\downarrow$} & original & dire\text{$\downarrow$}
& original & dire\text{$\downarrow$} & original & dire\text{$\downarrow$} \\
\hline
BERT & 74.2 & 47.5 & 94.1 & 73.9 & 83.8 & 64.1 & 71.0 & 36.5 & 64.2 & 32.5 \\
\quad+ours & 74.0 & \textbf{45.5} & \textbf{95.4} & \textbf{67.5} & 82.8 & \textbf{54.6} & \textbf{71.6} & \textbf{31.8} & 63.9 & \textbf{26.7} \\
\hline
XLNET & 76.2 & 50.3 & 96.5 & 75.0 & 86.6 & 64.8 & 74.4 & 39.1 & 68.0 & 34.6 \\
\quad+ours & 75.9 & \textbf{49.8} & \textbf{96.6} & \textbf{74.1} & 86.6 & \textbf{63.6} & 74.1 & \textbf{38.0} & 67.9 & \textbf{33.6} \\
\hline
DFGN & 71.7 & 44.5 & 94.4 & 73.8 & 83.8 & 64.0 & 68.7 & 34.2 & 62.1 & 30.4 \\
\quad+ours & \textbf{73.3} & 48.3 & \textbf{96.1} & \textbf{72.1} & \textbf{85.4} & \textbf{61.6} & 71.5 & 36.0 & 65.1 & 31.7 \\
\hline
HGN & 73.3 & 47.0 & 91.1 & 67.4 & 81.4 & 59.0 & 68.3 & 33.6 & 62.0 & 30.2 \\
\quad+ours & 70.9 & \textbf{41.1} & \textbf{93.4} & 67.6 & \textbf{83.5} & \textbf{58.1} & 67.6 & \textbf{28.9} & 61.8 & \textbf{25.5} \\
\hline
\end{tabular}
\caption{F1 scores. The "original" denotes the model's performance on the development set of HotpotQA in the distractor setting, and the "dire" indicates that the model scores on the corresponding probing set, which measures how much disconnected reasoning the model can achieve. The smaller score of "dire" is better. We can see that the proposed method can reduce disconnected reasoning while maintaining the same accuracy on the original dataset.  
}
\label{tab:f1}
\end{table*}
\begin{table*}
\centering
\begin{tabular}{ccccccccccc}
\hline
 & \multicolumn{2}{c}{\textbf{$\rm Ans$}} & \multicolumn{2}{c}{\textbf{$\rm Supp_p$}} & \multicolumn{2}{c}{\textbf{$\rm Supp_s$}} & \multicolumn{2}{c}{\textbf{$\rm Ans+Supp_p$}} & \multicolumn{2}{c}{\textbf{$\rm Ans+Supp_s$}}\\
& original & dire\text{$\downarrow$} & original & dire\text{$\downarrow$} & original & dire\text{$\downarrow$}
& original & dire\text{$\downarrow$} & original & dire\text{$\downarrow$} \\
\hline
BERT & 59.7 & 35.3 & 86.1 & 17.3 & 54.8 & 10.1 & 53.6 & 7.0 & 35.7 & 4.4 \\
\quad+ours & \textbf{60.2} & \textbf{33.7} & \textbf{87.9} & \textbf{3.7} & 53.5 & \textbf{2.0} & \textbf{55.2} & \textbf{1.4} & \textbf{36.6} & \textbf{0.8} \\
\hline
XLNET & 62.0 & 38.0 & 91.8 & 14.5 & 58.9 & 8.5 & 58.6 & 6.5 & 40.1 & 4.2 \\
\quad+ours & 61.7 & \textbf{37.6} & \textbf{92.1} & \textbf{6.9} & \textbf{59.3} & \textbf{3.9} & 58.5 & \textbf{2.6} & \textbf{40.9} & \textbf{1.6} \\
\hline
DFGN & 57.4 & 44.5 & 85.9 & 19.5 & 53.4 & 11.7 & 51.3 & 6.9 & 33.6 & 4.6 \\
\quad+ours & \textbf{59.6} & \textbf{35.9} & \textbf{90.9} & \textbf{6.2} & \textbf{57.7} & \textbf{3.3} & \textbf{55.9} & \textbf{2.3} & \textbf{38.4} & \textbf{1.4} \\
\hline
HGN & 58.9 & 35.2 & 79.5 & 16.9 & 52.3 & 10.4 & 49.5 & 6.8 & 34.2 & 4.4 \\
\quad+ours & 57.1 & \textbf{30.1} & \textbf{85.1} & \textbf{6.4} & \textbf{56.1} & \textbf{3.3} & \textbf{51.3} & \textbf{2.2} & \textbf{36.0} & \textbf{1.2} \\
\hline
\end{tabular}
\caption{EM scores on the development set and probing set of HotpotQA in the distractor setting.}
\label{tab:em}
\end{table*}

\textbf{Training}.  The model $\mathcal{F}$ is expected to disentangle the two natural direct effect and the true multi-hop effect from the total causal effect. To achieve this goal, we apply the Eq.~$(\ref{newY})$ to train the QA model $\mathcal{F}$. Our training strategy follows the HGN \citep{HGN}:
\begin{equation}
    \begin{aligned}
        \mathcal{L} &= \mathcal{L}_{start}+\mathcal{L}_{end}+\lambda\mathcal{L}_{sent}\\
        &+\mathcal{L}_{para}+\mathcal{L}_{type}+\mathcal{L}_{entity},
    \end{aligned}
\end{equation}
where $\lambda$ is a hyper-parameter and each term of $\mathcal{L}$ is cross-entropy loss function. 
Specifically, for answer prediction, we utilize the Eq. (\ref{newY}) to obtain the predicted logits of the start and end position of the answer span, and respectively calculate the cross-entropy loss $\mathcal{L}_{start}$ and $\mathcal{L}_{end}$ with corresponding ground truth labels. As for supporting facts prediction, similarly, we use the Eq. (\ref{newY}) to calculate the predicted logits in sentence level and paragraph level, and then calculate  $\mathcal{L}_{sent}$ and $\mathcal{L}_{para}$. We also apply our counterfactual reasoning method to identify the answer type \citep{DFGN,HGN}, which consists of yes, no, span and entity. We use the $[CLS]$ token as the global representation to predict the answer type under the Eq. (\ref{newY}) and calculate $\mathcal{L}_{type}$ with the ground truth label. Entity prediction ($\mathcal{L}_{entity}$) \citep{HGN} is only a regularization term and the Eq. (\ref{newY}) is not applied to this term.

\textbf{Inference}. As illustrated in Section \ref{Preliminaries}, our goal is to exclude the natural direct effect ($K_1\rightarrow Y, K_2\rightarrow Y$) and use the true multi-hop effect ($K\rightarrow Y$) to reduce the multi-hop QA model's disconnected reasoning, so we use Eq. (\ref{tie}) for inference:
\begin{equation}
     \mathcal{F}(q,s,c) - \mathcal{C}.
\end{equation}
The time consumption of our approach is equal to the existing methods. 

\section{Experiments}

We extensively conduct the experiments on the HotpotQA~\citep{hotpotqa} dataset. The compared results show that our proposed algorithm can reduce the amount of the disconnected reasoning while obtaining the strong performance on the original dev set. And it is general and suitable for other multi-hop QA architectures.

\textbf{Datasets:} We utilize the distractor setting of HotpotQA, where each question is equipped with two ground truth supporting paragraphs and eight distractor paragraphs as the context. And the answer of the question is annotated as a span in one of the supporting paragraphs. To measure  the disconnected reasoning (dire) in HotpotQA models, we generate the probing dataset following \citep{dire}, which is only used in the test phase. Specifically, the probing dataset for HotpotQA in the distractor setting divides each example of the original dataset into two instances, both of which only contain one of two ground truth supporting paragraphs respectively. If the multi-hop QA model can arrive at the correct test output on two instances, it means that the model performs disconnected reasoning on the original example. Please refer to \cite{dire} for more details. 

\textbf{Baselines:} First, we simply use the BERT~\citep{bert} to predict the answer, supporting sentences and supporting paragraphs as the baseline. $BERT+ours$ denotes that we apply our counterfactual multi-hop reasoning method based on BERT as the backbone. The proposed approach is model-agnostic and we also implement it on several multi-hop QA architectures, including DFGN~\citep{DFGN}, HGN~\citep{HGN} and XLNet in Dire \citep{dire,xlnet}. Our proposed algorithm also can be implemented on other baselines.

\begin{table*}
\centering
\begin{tabular}{ccccccccccc}
\hline
 & \multicolumn{2}{c}{\textbf{$\rm Ans$}} & \multicolumn{2}{c}{\textbf{$\rm Supp_p$}} & \multicolumn{2}{c}{\textbf{$\rm Supp_s$}} & \multicolumn{2}{c}{\textbf{$\rm Ans+Supp_p$}} & \multicolumn{2}{c}{\textbf{$\rm Ans+Supp_s$}}\\
& original & dire\text{$\downarrow$} & original & dire\text{$\downarrow$} & original & dire\text{$\downarrow$}
& original & dire\text{$\downarrow$} & original & dire\text{$\downarrow$} \\
\hline
BERT & 74.2 & 47.5 & 94.1 & 73.9 & 83.8 & 64.1 & 71.0 & 36.5 & 64.2 & 32.5 \\
+$K_1 \to Y$ & 74.2 & 50.2 & 96.6 & 73.5 & 85.7 & 61.9 & 72.5 & 37.8 & 65.7 & 32.8 \\
+$K_2 \to Y$ & 74.6 & 48.8 & 96.4 & 68.2 & 85.6 & 57.7 & 72.8 & 34.5 & 66.0 & 30.1 \\
+ours(full) & 74.0 & \textbf{45.5} & 95.4 & \textbf{67.5} & 82.8 & \textbf{54.6} & 71.6 & \textbf{31.8} & 63.9 & \textbf{26.7} \\
\hline
\end{tabular}
\caption{F1 scores of ablation study on the development set and probing set of HotpotQA in the Distractor setting. The "+ $K_1 \to Y$" denotes that we only utilize the counterfactual examples ($q,s,c^*$) to estimate the shortcut impact of $K_1 \to Y$, and similarly the "+ $K_2 \to Y$" represents that only the counterfactual examples ($q,s^*,c$) are used to estimate the shortcut impact of $K_2 \to Y$.}
\label{tab:ablation}
\end{table*}

\begin{table*}
\centering
\begin{tabular}{ccccccccccc}
\hline
 & \multicolumn{2}{c}{\textbf{$\rm Ans$}} & \multicolumn{2}{c}{\textbf{$\rm Supp_p$}} & \multicolumn{2}{c}{\textbf{$\rm Supp_s$}} & \multicolumn{2}{c}{\textbf{$\rm Ans+Supp_p$}} & \multicolumn{2}{c}{\textbf{$\rm Ans+Supp_s$}}\\
& original & dire\text{$\downarrow$} & original & dire\text{$\downarrow$} & original & dire\text{$\downarrow$}
& original & dire\text{$\downarrow$} & original & dire\text{$\downarrow$} \\
\hline
random & \textbf{74.0} & \textbf{45.5} & \textbf{95.4} & 67.5 & \textbf{82.8} & 54.6 & \textbf{71.6} & 31.8 & \textbf{63.9} & 26.7 \\
uniform & 73.8 & 45.9 & 92.1 & \textbf{67.4} & 76.6 & \textbf{46.3} & 69.2 & 31.8 & 59.6 & \textbf{23.2} \\
\hline
\end{tabular}
\caption{F1 scores of ablation study of assumptions for counterfactual outputs $\mathcal{C}$ on the development set and probing set of HotpotQA in the distractor setting.}
\label{tab:ablation_distri}
\end{table*}

\textbf{Metrics:} Following the Dire \citep{dire}, we report the metrics for HotpotQA: answer span ($\rm Ans$), supporting paragraphs ($\rm Supp_{p}$), supporting sentences ($\rm Supp_s$), joint metrics ($\rm Ans+Supp_p, Ans+Supp_s$). We show both EM scores and F1 scores to compare the performance between baselines and our counterfactual multi-hop reasoning method.

\subsection{Quantitative Results}
For fairness, we conduct the experiments under the same preprocessing of the dataset following HGN \citep{HGN}, which select top K relevant paragraphs corresponding to each question example. And the experimental results are shown in Table~\ref{tab:f1} and Table~\ref{tab:em}. The main observation can be made as follows:


\textit{Our method can reduce disconnected reasoning.} Compared to BERT baseline, our proposed counterfactual multi-hop reasoning method can reduce the disconnected reasoning of answer prediction and supporting facts identification in both the paragraph level and sentence level. In particular, we can see big drops of 9.5 F1 points on $\rm Supp_s$ (from 64.1 to 54.6) and 13.6 EM points on $\rm Supp_p$ (from 17.3 to 3.7) in disconnected reasoning (dire). Our method are better at reducing disconnected reasoning on the Exact Match (EM) evaluation metric. This is because EM is stricter evaluation metric. For example, EM requires both of the supporting facts should be predicted correctly while it has F1 scores even when only one supporting fact is predicted correctly. For dire evaluation where only one supporting fact is provided, our approach punishes this situation and achieves lower scores on EM metric of disconnected reasoning. It demonstrates that our method effectively reduce disconnected reasoning when there are without sufficient supporting facts. 

\textit{Our method still guarantees the comparable performance on the original dev set}. As seen from Table~\ref{tab:f1} and Table~\ref{tab:em}, the proposed method also maintain the same accuracy on the original set. It even shows the better performance on the supporting facts prediction in the paragraph level. 

\textit{Our method is model-agnostic and it demonstrates effectiveness in several multi-hop QA models. } Based on our causal-effect insight, our proposed approach can easily be applied to other multi-hop QA architectures including XLNET, DFGN, HGN \citep{dire,DFGN,HGN}. As shown in Table~\ref{tab:f1} and Table~\ref{tab:em}, our proposed counterfactual reasoning method achieves better performance. Our method can reduce the disconnected reasoning by introducing the proposed counterfactual approach in the training procedure.  The dire scores of HGN~\citep{HGN} and XLNET~\citep{dire} in $\rm Ans,Supp_p,Supp_s$ all drop to some extent. Besides, the performances on the original dev set are comparable simultaneously. It indicates that our proposed method is more stable and general. 

\begin{figure}
    \centering
    \includegraphics[width=7cm]{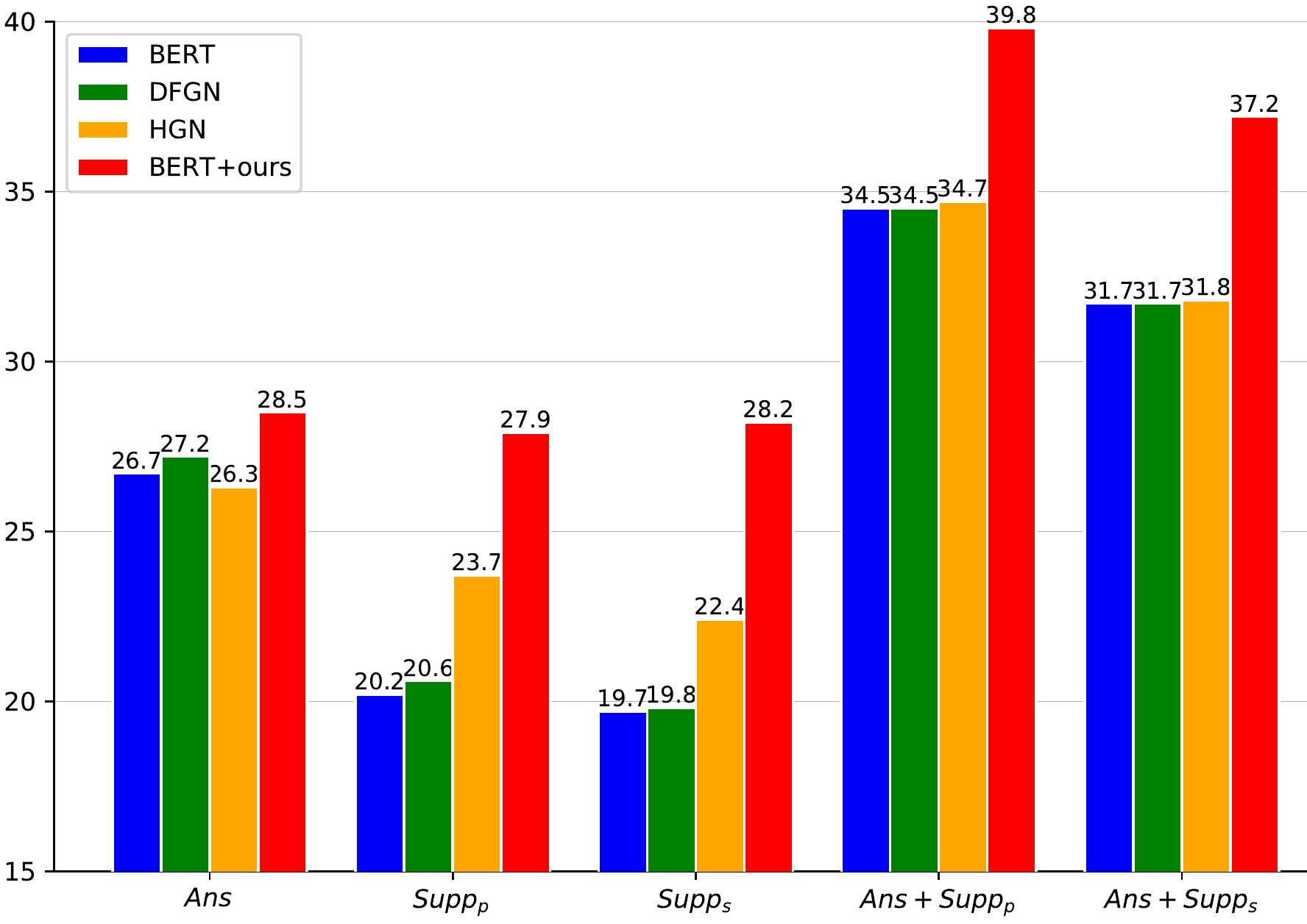}
    \caption{F1 scores of real multi-hop reasoning, which is denoted as the original scores minus the dire scores. We compare BERT, DFGN, HGN and ours method except XLNET, as they utilize the same pretrained language model (i.e. bert-base-uncased).}
    \label{fig:realmhp}
\end{figure}

In summary, reducing the disconnected reasoning and guarantee the strong performance on original development set indicate that the most progress of the model is attributed to the multi-top reasoning ($K \to Y$) capability. For intuitiveness, we also show the real multi-hop reasoning promoted by our proposed counterfactual reasoning approach, as shown in Fig.~\ref{fig:realmhp}. 

\subsection{Ablation Study}
As illustrated in Section \ref{implementation}, our goal is to exclude the shortcut impacts ($K_1 \to Y,K_2 \to Y$) to reduce the disconnected reasoning. Hence, we study the ablation experiments of excluding one of shortcuts impacts. We explore to remove $K_1 \to Y$ or  $K_2 \to Y$ that reduce the disconnected reasoning, as shown in Table \ref{tab:ablation}. We can see that exclude one of them can decrease the amount of disconnected reasoning to some extent on supporting facts identification except the answer span prediction. However, relieving the both impacts of $K_1 \to Y$ and $K_2 \to Y$ can achieve better performance on decreasing disconnected reasoning. Because the model can always exploit another shortcut if only one of the shortcuts is blocked.

We further conduct ablation studies to validate the distribution assumption for the counterfactual output of the parameter $\mathcal{C}$. Similar to CF-VQA \citep{counterfactualVQA}, we empirically validate the two distribution assumptions, as shown in Table~\ref{tab:ablation_distri}. The "random" denotes that $\mathcal{C}$ are learned without constraint and it means that $\mathcal{C}_{Ans}\in \mathbb{R}^n,\mathcal{C}_{supp}\in \mathbb{R}^2,\mathcal{C}_{type}\in \mathbb{R}^4$ respectively, and $n$ represents the length of the context. The "uniform" denotes that $\mathcal{C}$ should satisfy uniform distribution and it means that $\mathcal{C}_{Ans},\mathcal{C}_{supp}$ and $\mathcal{C}_{type}$ are scalar. As shown in Table \ref{tab:ablation_distri}, the random distribution assumption performs better than the uniform distribution assumption.

\section{Conclusion}
In this work, we proposed a novel counterfactual reasoning approach to reduce the disconnected reasoning in multi-hop QA. We used the causal graph to explain the existing multi-hop QA approaches' behaviors, which consists of the shortcut impacts and reasoning impacts. The shortcut impacts capture the disconnected reasoning. Thus the disconnected reasoning is formulated as natural direct causal effect and we construct the counterfactual examples during the training phase to estimate the both natural direct effects of question and context on answer prediction as well as supporting facts identification. The reasoning impact represents the multi-hop reasoning and is estimated via introducing learnable parameters.

During the test phase, we excluded the natural direct effect and utilize the true multi-hop effect to decrease the disconnected reasoning. Experimental results demonstrate that our proposed counterfactual reasoning method can significantly drop the disconnected reasoning on probing dataset and guarantee the strong performance on original dataset, which indicates the most profress of the multi-hop QA model is attributed to the true multi-hop reasoning.  Besides, our approach is model-agnostic, and can be applied to other multi-hop QA architectures to prevent the model from exploiting the shortcuts. 

Overall, our insights of reducing disconnected reasoning and learning the true multi-hop reasoning can motivate the development of new methods which contributes to promoting the multi-hop question answering task.

\section*{Limitations}
Our proposed method needs to construct counterfactual examples to estimate the natural direct effect of disconnected reasoning during the training phase, thus we need a little more GPU resources and computational time. However, the need of resource occupancy and time consumption of our approach does not increase during inference. Another limitation is that we use the learnable parameters to approximate the $Y_{k_1,k_2,k^*}$. In our future work, we will explore better approach to model it.

\bibliography{anthology,custom}
\bibliographystyle{acl_natbib}

\appendix

\section{Appendix}
\label{sec:appendix}

\subsection{Implementation Details}
\label{implementaionDetails}
Specifically, given the question $Q=q$, and context $P=\{s_1,s_2,...,s_m\}$ where $m$ is the number of paragraphs, we denote the remaining context $c_i=P-\{s_i\}$, $1\leq i \leq m$. To distinguish whether $s_i$ is supporting fact and get the answer distribution on $s_i$, we construct the $s_i^*$ and $c_i^*$ as illustrated in the subsection \ref{estimationOfeffect}. We respectively encode $(q,s_i,c_i)$,$(q,s_i^*,c_i)$ and $(q,s_i,c_i^*)$ to get the  contextualized representation $O\in R^{n\times d},M_i\in ^{n\times d},G_i \in R^{n\times d}$, where $n$ is the length of the question and context.

For supporting facts identification, similar to Dire \citep{dire}, we use the start token $s_i^{start}$ of the paragraph as its representation and obtain their predicted logits under factual and counterfactual scenario:
\begin{equation}
    \label{spY}
    \begin{aligned}
        &Y_{q,s_i,c_i}(s_i) = g(O[s_i^{start}]) \\
        &Y_{q,s_i^*,c_i}(s_i) = g(M_i[s_i^{start}]) \\
        &Y_{q,s_i,c_i^*}(s_i) = g(G_i[s_i^{start}]),
    \end{aligned}
\end{equation}
where $g$ is a classifier instantiated as $MLP$ layer in practice. And $s_i^{start}$ and $s_i^{end}$ are denoted as the start and end position of the paragraph $s_i$ respectively. In sentence level, we use the their start positions in paragraph $s_i$ and operate in the same way.

As for answer span prediction, we concatenate the representation of $s_i$ in $M_i$ or $G_i$ $(1\leq i \leq m)$ to construct the entire answer span prediction on the whole context:
\begin{equation}
    \label{ansY}
    \begin{aligned}
        &\bar{M}=[M_1[s_1^{start}:s_1^{end}];...;M_m[s_m^{start}:s_m^{end}]] \\
        &\bar{G}=[G_1[s_1^{start}:s_1^{end}];...;G_m[s_m^{start}:s_m^{end}]] \\
        &Y_{q,s,c}(start)=g(O)  \\
        &Y_{q,s^*,c}(start)=g(\bar{M})  \\
        &Y_{q,s,c^*}(start)=g(\bar{G}),
    \end{aligned}
\end{equation}
where $[;]$ denotes the operation of concatenation. And we can predict the end position of the answer in the same way.

We also apply our counterfactual reasoning method to identify the answer type, consisting of yes, no, span and entity:
\begin{equation}
    \begin{aligned}
        &Y_{q,s,c}(type)=g(O[0]) \\
        &Y_{q,s^*,c}(type)=\sum_{i=1}^{m} g(M_{i}[0]) \\
        &Y_{q,s,c^*}(type)=\sum_{i=1}^{m} g(G_{i}[0]).
    \end{aligned}
\end{equation}
We use the $[CLS]$ token as the global representation to predict the answer type, following previous work \citep{DFGN,HGN}.

\end{document}